\documentclass[12pt]{article}
\pdfoutput=1

\usepackage[utf8]{inputenc}
\usepackage[T1]{fontenc}
\usepackage{amsmath, amsthm, amsfonts, amssymb}
\usepackage{graphicx}
\usepackage{geometry}
\usepackage{hyperref}
\usepackage{cite}
\usepackage{enumitem}
\usepackage{caption}
\usepackage{wrapfig}
\newtheorem{theorem}{Theorem}
\newtheorem{lemma}[theorem]{Lemma}

\newtheorem{definition}[theorem]{Definition}

\newtheoremstyle{examplestyle}
      {}
    {}
    {\normalfont}
    {}
    {\bfseries}
    {.}
    {.5em}
    {}  
\theoremstyle{examplestyle}
\newtheorem{example}{Example}

\newcommand{\R}{\mathbb{R}}
\newcommand{\Z}{\mathbb{Z}}
\newcommand{\argmax}{\mathrm{argmax}}
\newcommand{\ip}[2]{\left\langle #1, #2 \right\rangle}
\newcommand{\relu}{\mathrm{relu}}
\newcommand{\softmax}{\mathrm{softmax}}
\newcommand{\round}{\mathrm{round}}
\newcommand{\next}{\mathrm{next}}
\newcommand{\aarg}{\mathrm{arg}}

\graphicspath{{figures/}}

\title{Banach-Tarski Embeddings and Transformers}
\author{Joshua Maher}
\date{October 2023}

\begin{document}

\maketitle

\begin{abstract}
We introduce a new construction of embeddings of arbitrary recursive data structures into high dimensional vectors.  These embeddings provide an interpretable model for the latent state vectors of transformers.  We demonstrate that these embeddings can be decoded to the original data structure when the embedding dimension is sufficiently large.  This decoding algorithm has a natural implementation as a transformer.  We also show that these embedding vectors can be manipulated directly to perform computations on the underlying data without decoding.  As an example we present an algorithm that constructs the embedded parse tree of an embedded token sequence using only vector operations in embedding space. 
\end{abstract}

\section{Introduction}
Transformer models, as introduced by Vaswani et al. \cite{Vas17}, have led to significant advancements in various machine learning domains.  One notable feature of these models is that embeddings produced from the internal activations of the models can effectively capture high-level information about the model inputs, where similar model inputs produce correlated embedding vectors \cite{Devlin2019BERTPO}.

To understand and interpret how transformer models work, we need a theory of how data is represented and manipulated within these models.  To work towards such a theory, we consider the following questions:  
\begin{enumerate}
\item Can we explicitly construct  embeddings that represent complex data structures?
\item Is is possible to construct algorithms that use these embeddings as data representations?
\end{enumerate}

In this paper, we describe a family of embeddings of arbitrary recursive data structures to vectors in $\R^d$. These embeddings are designed so that many natural operations on data structures translate to linear operations on the embedding vectors.  We call this the “BT Embedding” because it was inspired by the proof of the famous Banach-Tarski paradox.

The BT embedding is constructed using random vectors and matrices. The construction requires only a schema for the data, and does not need training or optimization.  It has the property that similarities in data structures can be detected using linear operations (e.g. dot products) to compare embedding vectors. 

This embedding enables the design of transformer models that compute directly on the embedding vectors and that do not require custom indexing schemes to represent nested data structures.  These constructions demonstrate that transformers can directly process and transform recursive data structures encoded as BT embedding vectors.

The main technical result is that the BT embedding is invertible with high probability when the dimension of the embedding is sufficiently large.  At a high level, the method is recursive application of the Johnson-Lindenstrauss lemma and 1-nearest-neighbors with respect to dot product similarity.   The embedding dimension required to reversibly encode a data structure in a single vector is approximately linear in the size of structure.  We will show that this decoding algorithm has a natural implementation as a transformer model.  

To illustrate how BT encodings can be used directly in algorithms on recursive data structures, we construct an algorithm to parse a sequence of BT encoded tokens with a collection of BT encoded production rules.  The result is a BT encoded parse tree of the input.  Notable, the algorithm operates without decoding any of the vectors, does not need access to the full schema of the input data, and can be implemented as a transformer.

These results can be interpreted as an analogy between the theory of data structures and linear algebra:

\begin{tabular}{|c|c|}
\hline
Computer Science & Linear Algebra \\
\hline
Symbols, Atomic Data Structures & Random Vectors / JL Embeddings \\
Attributes, Fields of Data Structures & Random Orthogonal Matrices \\
Recursive, Tree-like Data Structures &  Banach-Tarski Encoding \\
Paths in Data Structures & Representations of Free Groups on Attributes \\
Algorithms on Recursive Data Types & Transformers \\
\hline
\end{tabular}

We have implemented the encoding, decoding, transformer model, and parsing algorithms described in this paper.  The code, tests, and experiments are available at \url{https://github.com/jtmaher/Embedding}.

\subsection{Related Work}
In the paper ``Random Features for Large-Scale Kernel Machines'' \cite{Rahimi2007RandomFF} Rahimi and Recht showed that randomized features can be a powerful tool for various machine learning problems.  This work, along with the classic embedding lemma of Johnson and Lindenstrauss \cite{JL84} led us to consider the construction of embeddings from random vectors.

The Banach-Tarski Theorem \cite{BanachSurLD} \cite{Wagon1985TheBP} showed that there is a decomposition of a ball in $\R^3$ into several subsets that can be isometrically assembled into two copies of the original ball.  The proof of this theorem involves constructing a free group $F_2$ in the orthogonal group $O_3$ \cite{Groot1956FreeSO}.  The natural self-similar structure of the Cayley graph of the free group on two generators is key in the proof.  The presence of these groups in all orthogonal groups $O_d$, $d>2$ \cite{Cowling1997RandomSO}, led to the hypothesis that such symmetry groups could be used to create representations of tree structures in high dimensional vector spaces.

In ``Attention is Turing Complete'' by Perez et al \cite{Perez21} it was shown that transformers with positional embedding are Turing complete.  This work has been extended with more practical constructions in \cite{Weiss2021ThinkingLT}, \cite{Giannou2023LoopedTA}. Our construction of transformers in this paper uses similar methods.

In ``Neural Turing Machines'' by Graves et al \cite{Graves2014NeuralTM} it was demonstrated that state machines with memory can be directly optimized to perform computational tasks.

``Universal Transformers'' by Dehghani et al \cite{Dehghani2018UniversalT} introduced the idea that powerful transformer models can be produced by stacking identical (recurrent) transformer blocks.  While the transformers in our paper are not recurrent, our constructions are produced by iterating layers in a similar way (this method is also used in \cite{Giannou2023LoopedTA}).

In the context of tree kernels \cite{Collins2001ConvolutionKF}, Zanzotto and Dell'Archiprete \cite{Zanzotto2012DistributedTK} constructed features of trees using random embeddings of node labels.  Further work \cite{Ferrone2015DecodingDT} shows under certain conditions, parse trees of sentences can be decoded from the original sentence and these features.  This approach uses a nonlinear, weighted construction to combine node embeddings.

Shiv and Quirk in \cite{Shiv2019NovelPE} have constructed a positional embedding for paths in regular trees that can be traversed by linear operators.  Yao et al \cite{Yao2021SelfAttentionNC} produced a direct construction of a parser transformer for certain languages.

Hyperbolic geometry of activations has been proposed in \cite{Glehre2018HyperbolicAN} as a natural way to embed trees in neural networks.

\subsection{Acknowledgments}
The author would like to thank Misha Belkin and Aurora Maher for many interesting conversations and key feedback related to this work.  The author thanks Zhongqiang Huang for pointing out an error in a previous version of this paper.

\section{The BT Embedding}
\subsection{Definitions}
A \emph{schema} is a pair of finite sets $(T, A)$ where $T$ is a set of \emph{tokens} and $A$ is a set of \emph{attributes}.

A \emph{data structure} with schema $(T, A)$ is a finite tree where each node has a label from $T$ and each edge has an attribute from $A$.

A schema is called \emph{reflexive} if $A\subset T$.  This allows us to use attributes as values, which allows self describing structures.  We will assume that all schemas below are reflexive. 

Let $\mathcal{T}_{(T,A)}$ denote the set of all finite trees with nodes labeled with elements of $T$ and edges labeled with elements of $A$, with the restriction that a node can have at most one branch labeled any particular element of $A$

We think of these labeled trees as a general model of nested data structures, and for this reason we will use \emph{data structure} as a synonym for a tree in $\mathcal{T}_{(T,A)}$. 

Each node of such a tree has a \emph{path}, which is the unique sequence of edge labels starting at the root and ending at the node in question.

\subsection{Examples}

\begin{figure}[h]
  \caption
  \centering
  \includegraphics[width=0.6\textwidth]{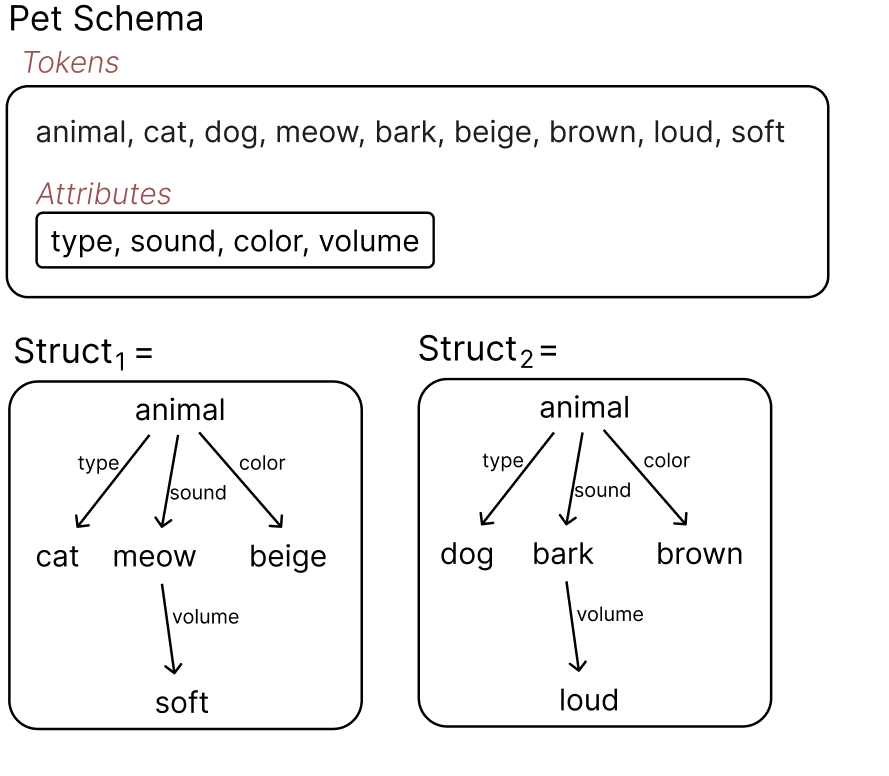}
  \label{fig:petschema}
\end{figure}

Figure \ref{fig:petschema} shows a simple schema for a data type that describes a pet.

This type of example can be generalized to any type of structured data, such as JSON, with a finite number of atomic values.  In practice, there are typically constraints on which attributes can be attached to a particular token, but in the present discussion we will ignore this detail. 

\begin{figure}[h]
  \caption
  \centering
  \includegraphics[width=0.6\textwidth]{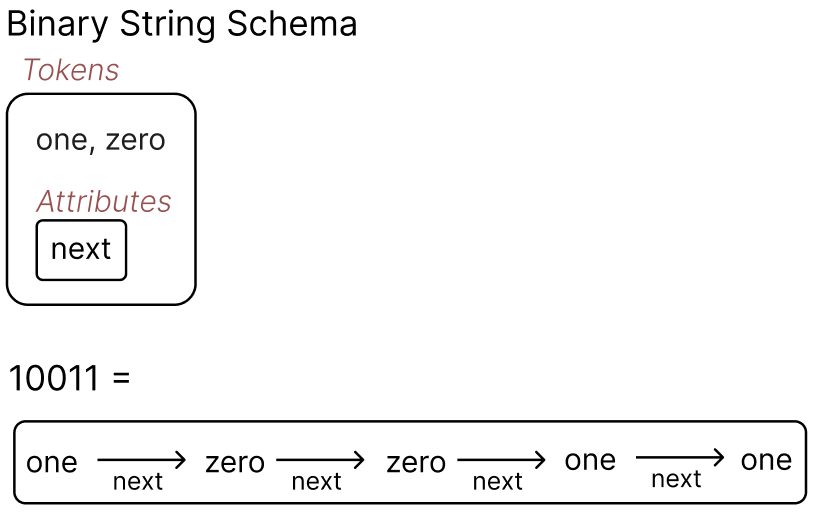}
  \label{fig:binaryschema}
\end{figure}

An important class of examples is linked lists, which can be modeled with a single “next” attribute (Figure \ref{fig:binaryschema}).   Further specializing to the case of binary lists allows representation of bit strings.

\subsection{Random Embeddings}
An \emph{embedding} $E$ of a schema $(T,A)$ is a pair of maps:

$$E_{token} : T \rightarrow \mathbb{S}^{d-1}\subset\mathbb{R}^d$$
$$E_{attr} : A \rightarrow O_d$$

In other words, each token is mapped to a unit vector, and each attribute is mapped to an orthogonal matrix.

Going forward, we will assume that $E$ is a \emph{random} embedding with respect to the uniform measure on $\mathbb{S}^{d-1}$ and Haar measure on $O_d$.  That said, it is possible that our constructions could be made substantially more efficient using other random embedding schemes, such as \cite{Achlioptas2003DatabasefriendlyRP}.

\subsection{BT Embedding}
The \emph{BT embedding} associated with $E$ is a map $BT_E:\mathcal{T}_{(T,A)}\rightarrow \mathbb{R}^d$ which extends $E$ to a map of all data structures to vectors.  Let $\tau\in\mathcal{T}_{(T,A)}$ be a tree:

$$BT_E(\tau) = \sum_{x \in nodes(\tau)} \Big( \prod_{i=1}^{\#(path(x))} E_{attr}(path(x)_i) \big) E_{tokens}(label(x))\Big)$$

In other words, the BT encoding of a tree is the sum of the embeddings of the tokens at the nodes, transformed by the product of matrices corresponding to the attributes of the path of each node.
Note that the order of operations in the matrix product is significant and runs from left to right as we move from the root to the leaves: i.e. $A_1 A_{2} \dots A_{k}$ where $A_i$ is the matrix corresponding to the $i$th element of the path.

We can extend this model by accepting trees with a \emph{set} of tokens for each node.  In this setting $BT$ is a linear map w.r.t. unions of trees.

Figure \ref{fig:petexample} shows how the embedding works for our pet schema example:

\begin{figure}[h]
  \caption
  \centering
  \includegraphics[width=0.6\textwidth]{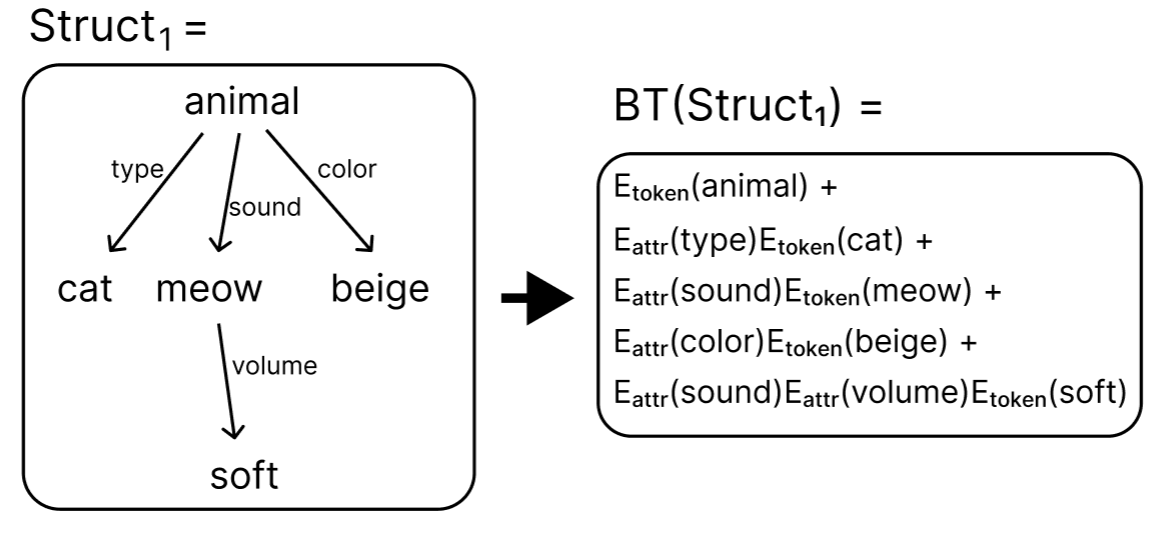}
  \label{fig:petexample}
\end{figure}

\subsection{Properties of the BT Embedding}
\subsubsection*{Cardinality}
$$\# nodes(x) \approx ||BT_E(x)||^2$$
This becomes arbitrary close to equality as the embedding dimension $d\rightarrow \infty$.

\subsubsection*{Linearity}
  Suppose that $\tau_1, \tau_2\in\mathcal{T}_{(T,A)}$ are trees, and $\nu\in Nodes(\tau_1)$ is a leaf node.  Given an attribute $a$, we can append $\tau_2$ to $\tau_1$ with the edge from $\nu$ to the root of $\tau_2$ labeled by $a$.  In this case we have:
$$BT(\tau_1 \cup_{\nu,a}\tau_2 ) = BT(\tau_1) + A \cdot BT(\tau_2)$$

where $A = ( E_{attr}(a)\prod_{p\in path(\nu)}E_{attr}(p))\in O_d$

This implies that if two trees share a common sub-tree, then there is a corresponding linear relationship between the BT embeddings.  Thus, the BT embedding gives us a way to transform arbitrary data into vectors that can plausibly be used for linear regression.

\subsubsection*{Extension}
If we extend BT to \emph{tree fragments}, where the node labels are optional (but where we still have a tree with all the edges labeled), then BT is linear on disjoint tree fragments.  While this extension can be useful, note that we cannot expect to decode fragments using a direct tree traversal, so decoding of fragments would likely be exponential time with respect to tree size.

\subsubsection*{Lists}
Linked lists, as in figure \ref{fig:binaryschema}, can be constructed in any schema using any attribute, but for clarity let us assume we have an attribute called ``next''.  For a list of tokens $t_1,\dots, t_n$, we can form a BT encoded list as follows, with the matrix $A_{\next}$ being the embedding of the ``next'' attribute:
$$\sum_{i=1,\dots,n} A_{\next}^{i-1}E(t_i)$$
This has analogous properties to the positional embeddings used in transformers \cite{Vas17}, namely, that translations of the original sequence of tokens correspond to multiplication by an orthogonal matrix.

We can push tokens onto an embedded list without decoding the list. If $v = BT({t_1,\dots,t_n})$ as above, then:
$$\mathrm{push}(v, t) = E(t) + A_{\next}\cdot v$$

On the other hand, popping tokens from the list requires decoding, and this illustrates a general property of BT embeddings:  \emph{Write} operations are linear, while \emph{read} operations are typically nonlinear, as we will see in the discussion of decoding.

\section{Decoding BT Embeddings}
In this section, we describe conditions under which the BT embedding is invertible. We introduce a recursive algorithm to decode these embeddings under these conditions.

Let $\tau \in \mathcal{T}_{(T, A)}$ be a data structure with $l$ nodes, and $v = BT_E(\tau)$ be its BT encoding with respect to randomly chosen $E_{tokens}\in \R^d, E_{attr}\in O_d$.  The following algorithm is based on the idea that we can find the most likely token for the root by maximizing the inner product between $v$ and the token vectors $E_{tokens}$.  To check for the presence of tokens associated with each attribute, we transform the vector $v$ by the inverse of the corresponding attribute matrix, and then attempt to decode a token in the same way.  Proceeding recursively we get:

\subsubsection*{Algorithm 1:}
\begin{enumerate}
\item Set $v = BT_E(\tau)$, output = empty tree, path = root.
\item For each $w_i \in E_{tokens}$, compute $x_i = \ip{w_i}{v}$.
\item If $\max{x_i} > 1/2$ then place a token $y$ corresponding to $\argmax(x_i)$ at the path on output, else return.
\item Transform $v$ by the inverse of each attribute matrix and recursively call step (2) with $v=A_i^{-1}v$ and path = path + $A_i$ for each $A_i$. 
\end{enumerate}

An implementation of this algorithm can be found at \url{https://github.com/jtmaher/Embedding/blob/master/embedding/encoder.py}. 

This algorithm runs in linear time with respect to the number of nodes in $\tau$, assuming that we can correctly decode all tokens (or lack thereof) with the comparison in step 3.  The tree structure is critical here, since if there is a disconnected tree fragment, we would need to search for tokens over an exponentially large set of potential nodes.

To understand how Algorithm 1 works, suppose the $y$ is the transformed embedding of the token at any node in $\tau$ and $v = BT(\tau)$.  We have:
\begin{equation}
  \ip{y}{v} = 1 + \sum_{i=1}^{l-1} \ip{y}{v_i}
  \label{eq1}
\end{equation}
where the $v_i$ range over all other $l-1$ terms in $BT(\tau)$.  Note that $l$ is the number of nodes of $\tau$.  If we happen to know that all the inner products $\ip{y}{v_i} < \epsilon$ for $v_i \neq y$ and where $\epsilon \cdot l < \frac{1}{2}$, then Algorithm 1 always picks the correct token  (or correctly infers that there is no token present if the maximum inner product is $< \frac{1}{2}$).

Let $\Gamma_l \subset O_d$ denote the set of all $l$-fold  products of the matrices in $E_{attr} \cup I_d$ where $I_d$ is the $d\times d$ identity.  Let $V = \Gamma_l E_{tokens}$ be the set of all products of the token vectors with these matrices.  $V$ is just the set of all possible terms in BT embeddings of trees with $\leq l$ nodes.  Combining this with the above analysis of the algorithm yields:

\begin{lemma}
If $|\ip{v_1}{v_2}| < \frac{1}{2l}$ for all $v_1 \neq v_2 \in V$, then Algorithm 1 decodes all trees with $\leq l$ nodes.
\label{bound1}
\end{lemma}

Intuitively, we expect that this condition is satisfied in sufficiently high dimensions, because the JL Lemma \cite{JL84} shows that the inner products of random vectors become increasingly clustered around zero as dimensionality is increased.

In fact, a looser bound than Lemma \ref{bound1} will suffice due to the fact that the terms in the sum of equation \ref{eq1} are essentially random and therefore combine sub-additively.  For instance, if we assume that the inner products $\ip{v_i}{v_j}$ are I.I.D., bounded in absolute value $|\ip{v_i}{v_j}|<\epsilon$, and mean zero for $i \neq j$, then we can conclude, by the central limit theorem, that we can pick $C$ such that the sums $S_l = \sum_{i=1}^{l-1} \ip{y}{v_i}$ in equation \ref{eq1} satisfy:
$$|S_l| < C \sqrt{l} \cdot \epsilon$$ 
with high probability as $l\rightarrow\infty$.  This yields a bound on the inner products that scales like $\frac{1}{\sqrt{l}}$ w.r.t. the tree size $l$:

\begin{lemma}
\label{bound2}
There is a constant $C$ such that if $\ip{v_i}{v_j}$ are I.I.D., $E(\ip{v_i}{v_j})=0$ and $|\ip{v_i}{v_j}| <  \frac{1}{2 C \sqrt{l}}$, then $S_l < \frac{1}{2}$ with high probability as $l\rightarrow\infty$.  Therefore Algorithm 1 succeeds with high probability under these assumptions. 
\end{lemma}

Note that we can control the bound on $|\ip{v_i}{v_j}|$ by increasing the embedding dimension $d$: According to the JL Lemma, the size of these inner products is proportional to $\frac{1}{\sqrt{d}}$.  This implies a linear relation between the embedding dimension and the size of trees that can be successfully decoded.  For a more detailed discussion, see Section \ref{JLdisc}.

\subsection{Empirical Decoding Results}

\subsubsection*{Random Lists}
\begin{figure}[h]
  \caption{Algorithm 1 Performance for Lists}
  \centering
  \includegraphics[width=0.8\textwidth]{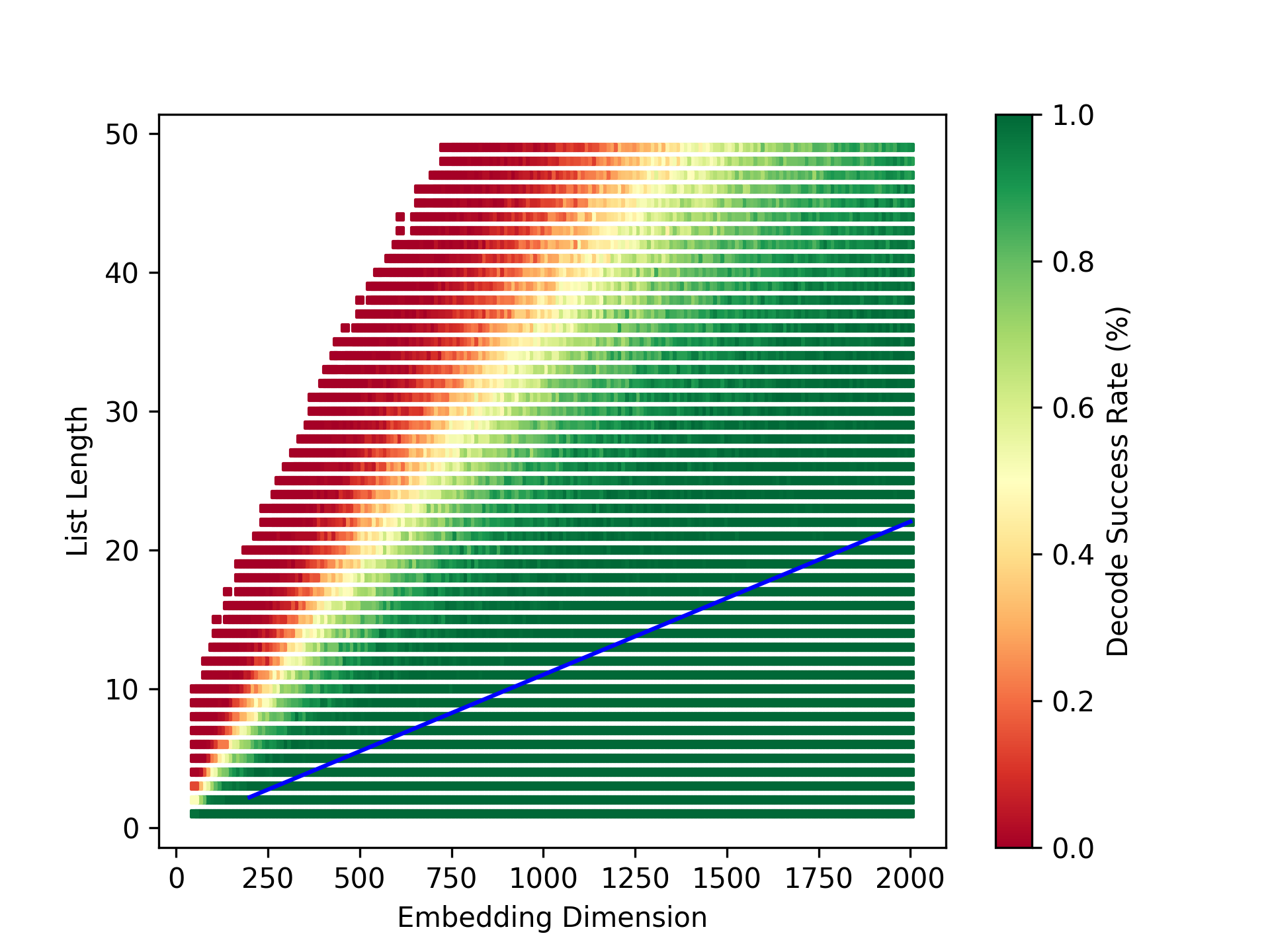}
  \label{fig:stressplot}
  \begin{minipage}{0.8\linewidth}
    \footnotesize
    All points below the blue line are at 100\% success rate
  \end{minipage}  
\end{figure}

Figure \ref{fig:stressplot} illustrates the results of an experiment on a schema for linked lists of 100 tokens with one attribute ``next''. For a selection of embedding dimensions and list lengths, we constructed 829,990 random lists and BT encoded them into vectors of various embedding dimensions.  The points on the graph indicate the success rate of decoding these vectors to the original lists.

The code for this experiment can be found at \url{https://github.com/jtmaher/Embedding/blob/master/Arrays.ipynb}.

\subsubsection*{Random Trees}
\begin{figure}[h]
  \caption{Algorithm 1 Performance for Trees}
  \centering
  \includegraphics[width=0.8\textwidth]{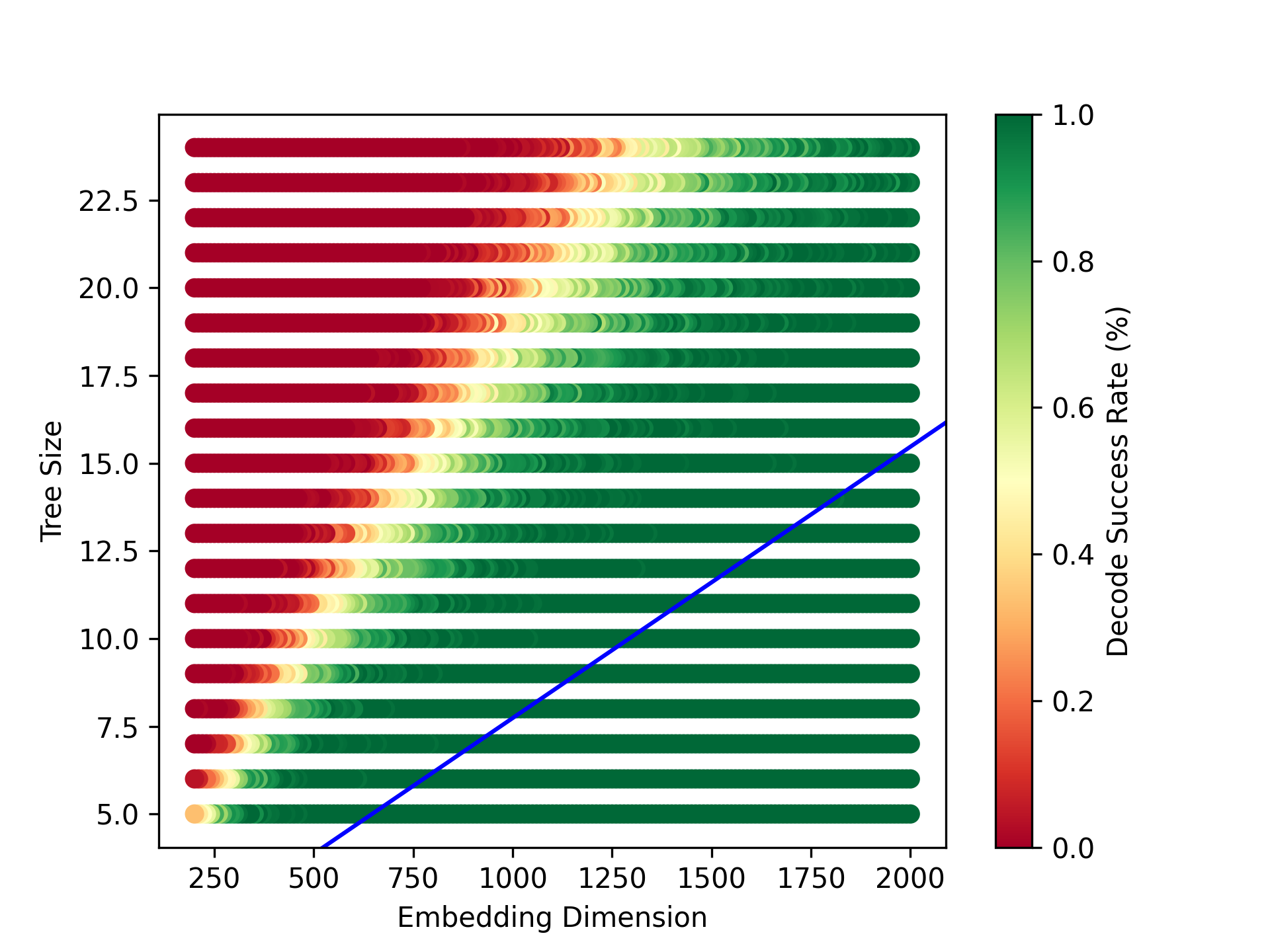}
  \label{fig:randplot}
  
  \begin{minipage}{0.8\linewidth}
    \footnotesize
    All points below the blue line are at 100\% success rate
  \end{minipage}

\end{figure}

Figure \ref{fig:randplot} shows the results of an experiment on a schema of 100 tokens and 4 attributes.  We randomly produced 196,000 embeddings of trees of size between 5 and 25 nodes, for embedding dimension $d \in [50,2000]$, and recorded the success rate by size and dimension. 

The code for this experiment can be found at \url{https://github.com/jtmaher/Embedding/blob/master/Random Trees.ipynb}.

\subsubsection*{Approximate Linear Scaling}
In both of these experiments, we observe that to successfully decode with probability approaching 1, we require approximately $d \gtrsim 125 \cdot l$ for embedding dimension $d$ and tree size $l$.

While this seems to require a very large number of dimensions, note that the set of structures of size $l$ is exponential in $l$ in both of these experiments, so a very large set of structures is being distinguished by this method.  In modern transformer models, the embedding space is indeed quite large, e.g. GPT-3 uses 12,288 dimensional embeddings \cite{Brown2020LanguageMA}, which is sufficient to embed structures of size $\approx 100$ using BT embeddings.

\section{Decoding with Transformers}
\label{transformer}
In this section, we describe how to implement the BT decoding algorithm with an explicit transformer model.  Our implementation uses similar techniques to \cite{Giannou2023LoopedTA}, \cite{Perez21}, and \cite{Yao2021SelfAttentionNC}.

Rather than implement the full recursive form of Algorithm 1, we will focus on the special case of retrieving the label of a node at a particular path from the BT encoding vector.  An implementation of this model can be found at \url{https://github.com/jtmaher/Embedding/blob/master/Transformers2.ipynb}.

\subsubsection*{Inputs and Outputs}

Let $\tau$ be a data structure.  The inputs to the transformer are:
\begin{enumerate}
\item A BT encoded vector $v = BT(\tau)$
\item A BT encoded path $r = BT([r_1,\dots,r_k])$, where $r_i$ are the tokens of the attributes in the desired path, and where the list is formed as a linked list using a distinguished ``next'' attribute.
\end{enumerate}

The transformer is a decoder-only (autoregressive) architecture and operates on sequences of $k$ vectors, where $k$ is the length of the path.  The input vector and path are placed in the first position of the sequence.

Recall that our schema is assumed to be reflexive, meaning that the attributes all have corresponding tokens, which is necessary to express paths in this way.
We also assume that we have a ``next'' attribute in the schema for formation of the path list.

The output of our model is a sequence of $k$ BT encoded tokens corresponding to the labels at each node along the path $p$, starting at the root.  The desired label at the input path will be found in the $k$-th position.  Although this is a ``decoder'' model, we note that the output and internal state will all consist of BT encoded data.  This feature was quite useful for debugging our implementation, since we could decode structures from the intermediate vectors to inspect the internal state of the model during each step of evaluation.

\subsubsection*{State Vectors}

We use position encoding vectors, which we define as follows: Let $p_1\in S^{k-1}\subset\R^k$ be a random unit vector and $Z\in O_k$ be random orthogonal matrix.  Then let $p_i =  Z^{i-1} t_1$.  We assume that the dimension $k$ is selected so that $\ip{p_i}{p_j}$ is small for $i\neq j \in [1,\dots,n]$, where $n$ is our sequence length.

\begin{figure}[h]
  \caption{Transformer Data Layout}
  \centering
  \includegraphics[width=0.8\textwidth]{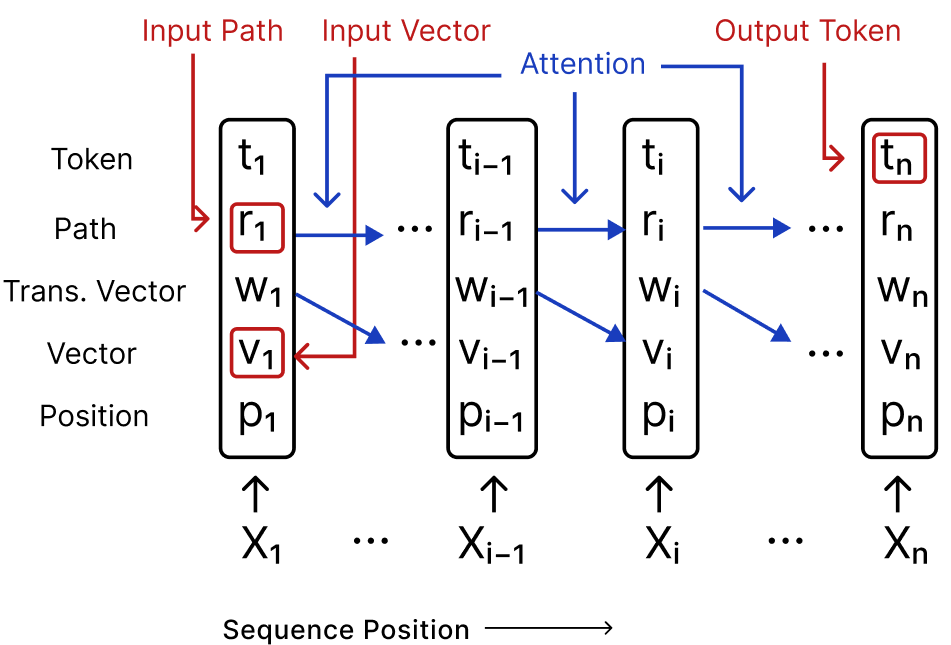}
  \label{fig:xformer1}

\end{figure}

Figure \ref{fig:xformer1} shows the data layout for our model.

For each position $i\in [1,\dots,n]$ in our sequence, we operate on a vector:
$$x_i = (p_i, v_i, w_i, r_i, t_i)$$
where $p_i$ is the position embedding, and $v_i,w_i, u_i, t_i\in \R^d$ are vectors in the $d$ dimensional BT embedding space.  Intuitively, $v_i$ is an ``input vector'', $w_i$ is a ``transformed vector'', $r_i$ is a ``path'', and $t_i$ is a ``token''.  The transformed vector will end up being the product of the input vector with the inverse of the attribute matrix corresponding to the current path entry.  The token will be decoded from the transformed matrix.

To initialize these vectors, we set:
$$x_1 = (p_1, v, 0, M_{\next} r, 0)$$
where $v$ is the input vector, and $r$ is the input path, and $M_{\next}$ is the BT attribute matrix corresponding to ``next''.  Multiplication by $M_{\next}$  ``shifts'' the path vector to skip the first position so that the root token can be decoded.

For $i>1$ set:
$$x_i = (p_i, 0, 0, 0, 0)$$

\subsubsection*{Attention Head}
We need to copy both the transformed vector and the path from each position in the sequence to the next.  This is accomplished with a single attention head.

Define $A$ to be the causal attention block on our sequence of vectors $x_i$, where the query, key and value are:
$$Q(x_i) = Z^{-1} p_i$$
$$K(x_i) = p_i$$
$$V(x_i) = (0, w_i, 0, M_{\next}^{-1} r_i, 0)$$

For sufficiently low softmax temperature, $A$ mostly attends to the $i-1$th position and produces the transformed token $w_{i-1}$ and a shifted copy of the path vector (except for the degenerate case of $i=1$, where we retrieve the zero vector by definition).

In other words, if $i>1$ we have:
$$x_i + A(x_i) = x_i + \sum_{j<i} \softmax_j(\ip{Z^{-1} p_i}{p_j}) (0,w_j,0,M_{\next}^{-1}  r_j,0)$$  
$$\approx (p_i,w_{i-1},w_i,M_{\next}^{-1} r_{i-1},t_i)$$
Note that the transformed vector $w_i$ is being moved to the $v_i$ position in the output.

\subsubsection*{Feed Forward Layers}
We construct two feed forward layers of the form $F(x_i) = M_1 \relu (M_2 x_i) + x_i$ where $M_1, M_2$ are affine transformations (i.e. linear with a constant bias).

First, we need to transform the input vector $v_i$ by the inverse of the attribute matrix corresponding to the attribute token at the front of the path $r_i$. Let $E_{attr} \subset E$ denote the token embeddings of the attributes and all tokens, respectively. 

Let:
$$y = C (E_{attr} r_i - \frac{1}{2}) \in \R^k$$
where $C$ is a large constant and $k$ is the number of attributes in the schema. $y$ is intended to be very positive on the dimension corresponding to the path attribute encoded at the head of $r$, and very negative otherwise.

We can use $y$ to form a conditionally transformed version of $v$ using a technique from \cite{Giannou2023LoopedTA}:
$$f_1(x_i) = \relu(v_i)-\relu(-v_i) + \sum_j \relu(y_i + M_j^{-1} v_i - v_i) - \relu(y_i)$$
where $M_j$ is the $j$th attribute matrix.  It is easy to see that $f_1(x_i) = M^{-1} v_i$ where $M$ is the attribute matrix corresponding to the head of the path vector, or if there is no token decoded from the path vector (i.e. $i=0$), $f_1(x_i) = v_i$.

Now we can define the first feed forward layer:
$$F_1(x_i) = (0, 0, f_1(x_i) - \relu(w_i) + \relu(-w_i), 0, 0) + x_i$$

The additional relu terms here remove the residual contribution of $w_i$.

At this point, we need to decode the token at the root of the transformed vector.  Let:
$$z = C (E w_i - \frac{1}{2})$$
$$f_2(x_i) = E^{T}(\mathrm{relu}(z + 1) - \mathrm{relu}(z))$$ 
where $C$ is a large constant.  The purpose of $z$ is to pick out the token with the largest dot product with $w_i$, which is then converted to the relevant token embedding by the relu expression.
Now let:
$$F_2(x_i) = (0, -\relu(v_i) +\relu(-v_i), 0, 0, f_2(x_i)) + x_i$$
The additional relu terms in the $v_i$ component are to zero out this term when combined with the residual state vector $x_i$.

\subsubsection*{Decoding Block}

To complete the construction, we put everything together into a decoding ``block'':
$$D(x_i) = F_2(F_1(A(x_i)+x_i))$$

We can think of $D$ as propagating the transformation of the input and path one step forward in the sequence, while simultaneously decoding tokens.

The full transformer decoding model is simply the $n-1$-fold iteration of the $D$ block:
$$x^{out}= (p^{out}_i, v^{out}_i, w^{out}_i, r^{out}_i, t^{out}_i) = D^{n-1}(x_i)$$
Assuming that the input path has length $n-1$, the output token corresponding to the input path is $t^{out}_n$.  The intermediate tokens of $v$ are decoded to $t^{out}_i$ for $i \in [1, \dots, n]$. 

Because this decoding algorithm uses the same inner product logic as Algorithm 1, it will work under identical conditions.

Note that we have two feed forward layers per attention layer, whereas the usual transformers have only one - this is easily remedied by adding a zero-valued attention layer between $F_1$ and $F_2$ if desired.

\section{Parsing with BT Embeddings}
In this section we present an algorithm to parse a sequence of BT encoded tokens according to a finite set of BT encoded production rules.  This algorithm does not require decoding of any of the structures and produces the parse tree as a BT encoded vector.  In fact, it does not even need access to the schema of the input, except for a small number of attribute matrices required to build up the output tree.

This algorithm can be implemented as a transformer in a similar way as the decoder transformer in Section \ref{transformer}.

We consider the problem of producing the parse tree of a sequence of $n$ tokens, according to $k$ $m$-ary production rules (i.e. the rules match up to $m$ tokens).  

We need a distinguished attribute $\next$ to build lists, and $m$ distinguished attributes $\aarg_1, \dots, \aarg_m$ that will be used to store the children of replacement node  to form the parse tree.  Assume we are using a schema $(T,A)$ that includes these attributes.

Each production rule consists of a \emph{pattern} $P_i$ and a \emph{replacement} $R_i$.  Both of these are BT-embedding vectors in $\R^d$.

The $P_i$ will be the encoding of a sequence of $\leq m$ tokens, using the $\next$ attribute linked list construction.  We will assume here that $R_i$ is a single token indicating the expression type of the replacement value of the production rule.

\begin{example}
The language of balanced parentheses, where we set $L = $ `(' and $R =$ `)'.  The tokens are $\{L,R,E\}$, where $E$ is a placeholder for “expression”.
\label{balanced}
\end{example}
\begin{tabular}{|c|c|c|c|}
\hline
Pattern & Replacement & $P_i\in \R^d$ & $R_i \in \R^d$ \\
\hline
 $LR$ & $E$ & $BT(L) + A_\next BT(R)$ & $BT(E)$ \\
 $LER$ & $E$ & $BT(L) + A_\next BT(E) + A_\next^2 BT(R)$& $BT(E)$ \\
 $E\ E$ & $E$& $BT(E) + A_\next BT(E)$ & $BT(E)$ \\
\hline
\end{tabular}

Since this language uses ternary productions, the attributes required for parsing are $\{\next, \aarg_1, \aarg_2,\aarg_3\}$.

\subsection{Algorithm 2}

Assume that we have a set of production rules expressed as strings of tokens in a schema $(T,A)$.  We will denote the BT encoded attribute matrices for our distinguished attributes as $A_{\next}, A_{\aarg_1},\dots$.

\begin{enumerate}
\item BT encode each pattern $P_i$ as a list $p_i = \sum_{0<j\leq len(p_i)} A_{\next}^{j-1} P_{i,j}$.
\item Let $r_i = BT(R_i)$ be the encoding of the replacements.
\item Let $x_i = BT(X_i)$ be the encoding of the input sequence.
\item For each $p_i$
  \begin{enumerate}
  \item Let $m = len(p_i)$.  Note that this can be computed as $\round(||p_i||^2)$ in sufficient large embedding dimension.
  \item For each consecutive $m$-tuple of inputs $x_j, \dots, x_{j+m}$, form a BT encoded list $x = \sum_{0<k\leq m} A_{\next}^{k-1} x_{j+k}$
    \begin{enumerate}
    \item Test for a match by comparing $\ip{p_i}{x} > m-1/2$.  If false, continue on the next $m$-tuple $x_{j+1},\dots$.
    \item If true, we replace the entire input $m-$tuple with the result $r_i$ corresponding to pattern $i$, adding the matched tokens as attributes:  $r_i + \sum_{0<k\leq m} A_{\aarg_k}x_{j+k}$
    \end{enumerate}
  \end{enumerate}
\item Repeat step (4) until no matches are found.
\end{enumerate}

Note that this algorithm does not require access to the full embedding of $(A,T)$, only the matrices $A_{\next}, A_{\aarg_i}$.

An implementation of this algorithm can be found at \url{https://github.com/jtmaher/Embedding/blob/master/embedding/parser.py}.

The key point is to determine when we can successfully test for a match with the comparison $\ip{p_i}{x} > m-1/2$.  Assuming that $p_i$ encodes a tree of size $m$, and $x$ is size $l$, expanding both arguments of $\ip{p_i}{x}$ into their BT encoding sums yields $m\cdot l$ terms.

Assuming that the non-matching terms have inner products $<\frac{1}{2C \sqrt{ml}}$, summing up to $ml$ terms, the result will be $< 1/2$ with high probability for an appropriate choice of the constant $C$.  So, by the same logic as Lemma \ref{bound2}, we get:

\begin{lemma}
  There is a constant $C$ such that if:
  \begin{enumerate}
  \item $\ip{v_i}{v_j}$ are I.I.D.
  \item $E(\ip{v_i}{v_j})=0$
  \item $|\ip{v_i}{v_j}| < \frac{1}{2 C \sqrt{ml}}$ for $i<j$
    \item $x_1, \dots, x_n$ is an input sequence of vectors, whose parse tree has $\leq l$ nodes, and the maximum arity of the patterns is $m$
  \end{enumerate}
  then Algorithm 2 produces the correct parse tree with high probability for large $m\cdot l$. 
\end{lemma}

Note that the intermediate $x_i$ terms are parse trees, and the size of these terms is not necessarily bounded in terms of the input length (for general production rules), which is why we bound in terms of the parse tree size, not the input length.

Observe that all the data at every step of the algorithm is BT encoded, which means that we can inspect or otherwise make use of the intermediate results.

\subsection{Empirical Parsing Results}

We implemented Algorithm 2 and ran it on an assortment of 17,185 randomly selected sequences of balanced parentheses of length $\leq 34$, with embedding dimensions from 200 to 2000.  The desired output is a BT encoded parse tree in the first (and only) slot of the output sequence.

In this test, we considered an output to be successful if and only if the output decoded to a correct parse tree of the original input using Algorithm 1.

As in the tests of decoding, we see an approximately linear scaling of the maximum successful sequence length with respect to embedding dimension.

\begin{figure}[h]
  \caption{Algorithm 2 Performance}
  \centering
  \includegraphics[width=0.8\textwidth]{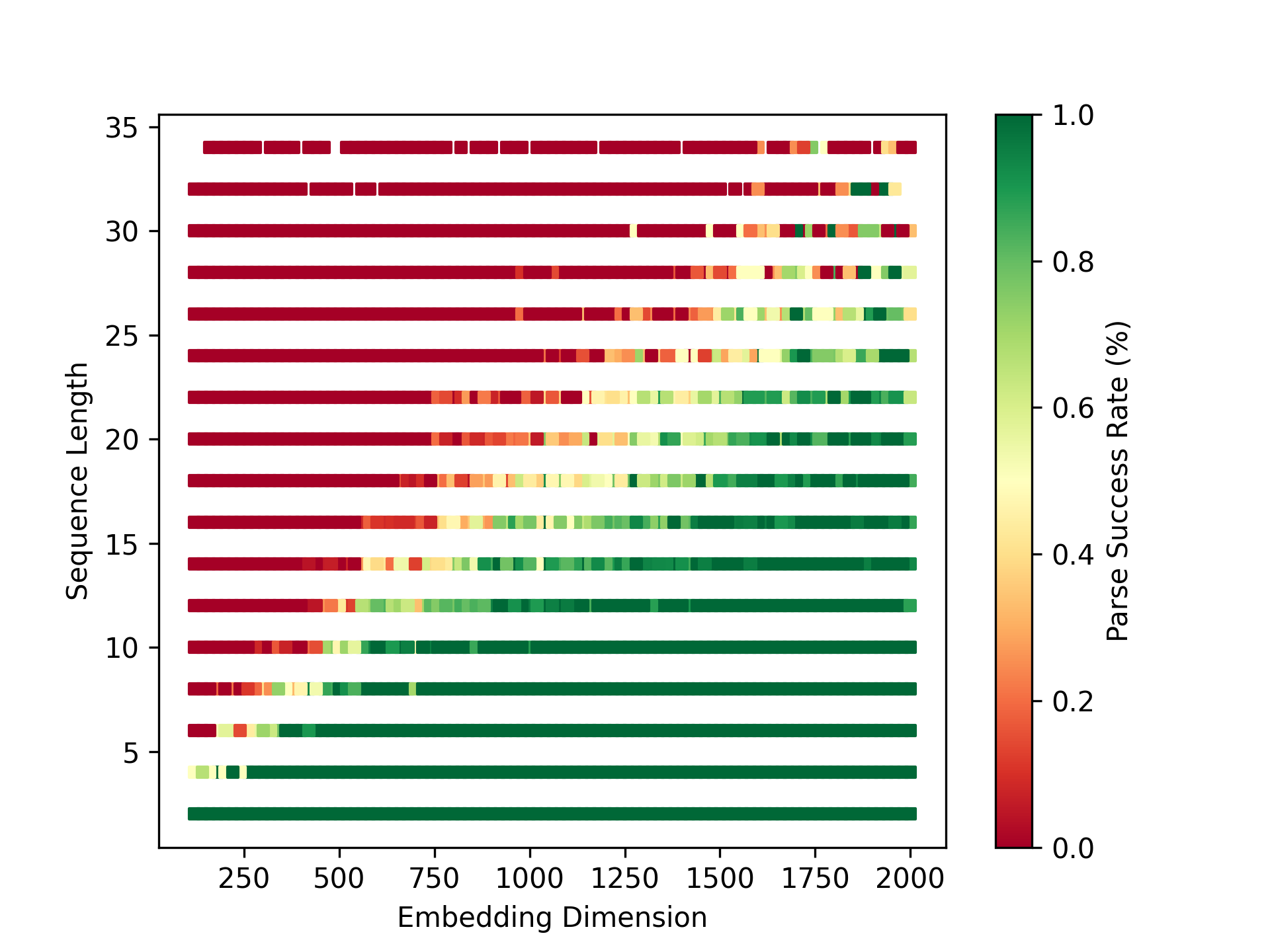}
  \label{fig:parse}
\end{figure}

The code for this experiment can be found at \url{https://github.com/jtmaher/Embedding/blob/master/Parser.ipynb}.

\section{The JL Lemma and Decoding}
\label{JLdisc}
A natural question is whether the conditions of Lemma \ref{bound2} hold for sufficiently high embedding dimension.  We will sketch an argument that the conditions do hold in the special case where our schema contains only one attribute.  The general case with $>1$ attributes appears to work in practice, but we do not have a proof.

First, we recall the JL Lemma \cite{JL84}, \cite{Dasgupta2003AnEP}, \cite{linial1995geometry}:
\begin{lemma}
  If we have a collection of $n$ independent uniformly distributed random unit vectors $V \subset S^{d-1}\subset \R^d$, then if $d > 16 \frac{log(n)}{\epsilon^2}$, then $|\ip{v_i}{v_j}| < \epsilon$ for any pair of vectors $v_i, v_j\in V$, with high probability.
\end{lemma}

To show that we can decode BT embeddings with Algorithm 1 in sufficiently high dimensions, we require a similar bound on inner products when $V$ is augmented with orthogonally transformed copies of itself, so that we can apply Lemma \ref{bound1} or Lemma \ref{bound2}.

\subsubsection*{Scaling Law}
Note that if the conclusion of JL did hold on this augmented set of vectors, then we could substitute $\epsilon$ with the bound from Lemma 2 and get:
$$ d > 16 \frac{log(n)}{\epsilon^2}$$
$$ d > 64  C^2 \cdot l \cdot log(n)$$
In Lemma \ref{bound2} we only need the conclusion of JL for sub-collections of $n=l\cdot T$ vectors, where $T$ is the number of tokens in the schema.  Therefore if $d > 64 C^2 l\cdot log(lT)$, we would expect to decode trees of size $\leq l$ with high probability.  Since this is approximately linear in $l$ for small $l$, it would agree with the empirical scaling results above.

\subsubsection*{Random Matrices}

Let $A_1, \dots, A_k \in O_d$ represent independent Haar uniform random orthogonal matrices.  Let $\Gamma < O_d$ denote the subgroup generated by the $A_i$.  Let $\Gamma V$ denote the set of all unit vectors of the form $g v$ for $g\in \Gamma, v \in V$.

\begin{definition}
We say that $\Gamma V$ \emph{$m$-$\epsilon$ separated} if
for all $v_1, v_2\in W \subset \Gamma V$ with $v_1\neq v_2$, we have $|\ip{v_1}{v_2}| < \epsilon$ with high probability, for all subsets $W\subset \Gamma V$ of cardinality $\leq m$.
\end{definition}

In other words, the condition is asserting that the conclusion of the JL Lemma continues to hold for subsets of $\Gamma V$.   

We would like to have a result that implies $m$-$\epsilon$ separation with high probability given $d > C \frac{log(m)}{\epsilon^2}$ for some constant $C$.  Some intuition comes from \cite{Cowling1997RandomSO} which proved that the subgroup of $SO_d$ generated by two or more random elements is a free and dense subgroup with probability one, which makes it plausible that the vectors in $\Gamma V$ are as evenly dispersed as if they were random. This is not true, as can be seen by examining the case of the powers of a single random orthogonal matrix $A$:

For a full measure set of $A\in O_d$, $\cup_{n=1,\dots,\infty} A^n$ is dense in a maximal torus of $O_d$, which has dimension $\lfloor d/2\rfloor$.  Thus, it is easy to see that the orbit of a vector $v \in S^{d-1}$ must be dense in a $\lfloor d/2\rfloor-1$ dimensional torus in $S^{d-1}$.  This can be seen explicitly by writing $A$ in block diagonal form.  Since $\Gamma$ acts ergodically on this torus, the distribution of the points in $\Gamma V$ are uniformly distributed (with respect to an appropriate limit).  Therefore we can conclude:

\begin{lemma}
If $\Gamma = \{A^i, i\in\Z \}$ and $V\subset S^{d-1}$ is an independent uniformly distributed set of unit vectors, then $\Gamma V$ is $m$-$\epsilon$ separated when $d > 32 \frac{log(m)}{\epsilon^2}$. (Note that the constant is twice that in the JL Lemma.)
 \end{lemma}
From this and Lemma \ref{bound2} we can conclude that there is a constant $C$ such that the BT embedding is invertible with high probability for schemas with one attribute and $T$ tokens, when the embedding dimension satisfies:
$$ d > 128 C^2 \cdot l \cdot log(l \cdot T)$$

Note that this bound on embedding dimension is nearly linear in the tree size $l$.

\newpage
\bibliographystyle{plain}
\bibliography{refs}

\begin{thebibliography}{10}

\bibitem{Achlioptas2003DatabasefriendlyRP}
Dimitris Achlioptas.
\newblock Database-friendly random projections: Johnson-lindenstrauss with
  binary coins.
\newblock {\em J. Comput. Syst. Sci.}, 66:671--687, 2003.

\bibitem{BanachSurLD}
Stefan Banach and Alfred Tarski.
\newblock Sur la d{\'e}composition des ensembles de points en parties
  respectivement congruentes.
\newblock {\em Fundamenta Mathematicae}, 6:244--277, 1924.

\bibitem{Brown2020LanguageMA}
Tom~B. Brown, Benjamin Mann, Nick Ryder, Melanie Subbiah, Jared Kaplan,
  Prafulla Dhariwal, Arvind Neelakantan, Pranav Shyam, Girish Sastry, Amanda
  Askell, Sandhini Agarwal, Ariel Herbert-Voss, Gretchen Krueger, T.~J.
  Henighan, Rewon Child, Aditya Ramesh, Daniel~M. Ziegler, Jeff Wu, Clemens
  Winter, Christopher Hesse, Mark Chen, Eric Sigler, Mateusz Litwin, Scott
  Gray, Benjamin Chess, Jack Clark, Christopher Berner, Sam McCandlish, Alec
  Radford, Ilya Sutskever, and Dario Amodei.
\newblock Language models are few-shot learners.
\newblock {\em ArXiv}, abs/2005.14165, 2020.

\bibitem{Collins2001ConvolutionKF}
Michael Collins and Nigel~P. Duffy.
\newblock Convolution kernels for natural language.
\newblock In {\em NIPS}, 2001.

\bibitem{Cowling1997RandomSO}
Michael~G. Cowling and Brian Dorofaeff.
\newblock Random subgroups of lie groups.
\newblock {\em Rendiconti del Seminario Matematico e Fisico di Milano},
  67:95--101, 1997.

\bibitem{Dasgupta2003AnEP}
Sanjoy Dasgupta and Anupam Gupta.
\newblock An elementary proof of a theorem of johnson and lindenstrauss.
\newblock {\em Random Structures \& Algorithms}, 22, 2003.

\bibitem{Groot1956FreeSO}
J.~de~Groot and T.~J. Dekker.
\newblock Free subgroups of the orthogonal group.
\newblock {\em Compositio Mathematica}, 12:134--136, 1956.

\bibitem{Dehghani2018UniversalT}
Mostafa Dehghani, Stephan Gouws, Oriol Vinyals, Jakob Uszkoreit, and Lukasz
  Kaiser.
\newblock Universal transformers.
\newblock {\em ArXiv}, abs/1807.03819, 2018.

\bibitem{Devlin2019BERTPO}
Jacob Devlin, Ming-Wei Chang, Kenton Lee, and Kristina Toutanova.
\newblock Bert: Pre-training of deep bidirectional transformers for language
  understanding.
\newblock In {\em North American Chapter of the Association for Computational
  Linguistics}, 2019.

\bibitem{Ferrone2015DecodingDT}
Lorenzo Ferrone, Fabio~Massimo Zanzotto, and Xavier Carreras.
\newblock Decoding distributed tree structures.
\newblock In {\em International Conference on Statistical Language and Speech
  Processing}, 2015.

\bibitem{Giannou2023LoopedTA}
Angeliki Giannou, Shashank Rajput, Jy~yong Sohn, Kangwook Lee, Jason~D. Lee,
  and Dimitris Papailiopoulos.
\newblock Looped transformers as programmable computers.
\newblock {\em ArXiv}, abs/2301.13196, 2023.

\bibitem{Graves2014NeuralTM}
Alex Graves, Greg Wayne, and Ivo Danihelka.
\newblock Neural turing machines.
\newblock {\em ArXiv}, abs/1410.5401, 2014.

\bibitem{JL84}
William~B. Johnson and Joram Lindenstrauss.
\newblock Extensions of lipschitz mappings into a hilbert space.
\newblock {\em Contemporary Mathematics}, 26:189--206, 1984.

\bibitem{linial1995geometry}
Nathan Linial, Elon London, and Yuri Rabinovich.
\newblock The geometry of graphs and some of its algorithmic applications.
\newblock {\em Combinatorica}, 15(2):215--245, 1995.

\bibitem{Perez21}
Jorge Pérez, Pablo Barceló, and Javier Marinkovic.
\newblock Attention is turing-complete.
\newblock {\em Journal of Machine Learning Research}, 22(75):1--35, 2021.

\bibitem{Rahimi2007RandomFF}
Ali Rahimi and Benjamin Recht.
\newblock Random features for large-scale kernel machines.
\newblock In {\em NIPS}, 2007.

\bibitem{Shiv2019NovelPE}
Vighnesh~Leonardo Shiv and Chris Quirk.
\newblock Novel positional encodings to enable tree-based transformers.
\newblock In {\em Neural Information Processing Systems}, 2019.

\bibitem{Vas17}
Ashish Vaswani, Noam Shazeer, Niki Parmar, Jakob Uszkoreit, Llion Jones,
  Aidan~N Gomez, \L~ukasz Kaiser, and Illia Polosukhin.
\newblock Attention is all you need.
\newblock In {\em Advances in Neural Information Processing Systems},
  volume~30, 2017.

\bibitem{Wagon1985TheBP}
Stan Wagon.
\newblock {\em The Banach-Tarski Paradox}.
\newblock Cambridge University Press, 1985.

\bibitem{Weiss2021ThinkingLT}
Gail Weiss, Yoav Goldberg, and Eran Yahav.
\newblock Thinking like transformers.
\newblock {\em ArXiv}, abs/2106.06981, 2021.

\bibitem{Yao2021SelfAttentionNC}
Shunyu Yao, Binghui Peng, Christos~H. Papadimitriou, and Karthik Narasimhan.
\newblock Self-attention networks can process bounded hierarchical languages.
\newblock In {\em Annual Meeting of the Association for Computational
  Linguistics}, 2021.

\bibitem{Zanzotto2012DistributedTK}
Fabio~Massimo Zanzotto and Lorenzo Dell'Arciprete.
\newblock Distributed tree kernels.
\newblock {\em ArXiv}, abs/1206.4607, 2012.

\bibitem{Glehre2018HyperbolicAN}
Çaglar G{\"u}lçehre, Misha Denil, Mateusz Malinowski, Ali Razavi, Razvan
  Pascanu, Karl~Moritz Hermann, Peter~W. Battaglia, Victor Bapst, David Raposo,
  Adam Santoro, and Nando de~Freitas.
\newblock Hyperbolic attention networks.
\newblock {\em ArXiv}, abs/1805.09786, 2018.

\end{thebibliography}

\end{document}